\documentclass[11pt]{article}
\usepackage[margin=1in]{geometry}
\usepackage{booktabs}
\usepackage{hyperref}
\usepackage[numbers]{natbib}
\usepackage{tikz}
\usetikzlibrary{positioning, arrows.meta, shapes.geometric, fit, backgrounds}
\usepackage{amsmath,amssymb}
\usepackage{graphicx}
\usepackage{xcolor}

\definecolor{pgblue}{HTML}{336699}
\definecolor{cfgreen}{HTML}{339966}
\definecolor{agentred}{HTML}{CC4444}
\definecolor{fileyellow}{HTML}{CC9933}
\definecolor{bglight}{HTML}{F5F5F5}

\title{PrimeAgentOrchestrator: Memory-Primed Agent Spawning\\for Personal AI Infrastructure}
\author{Myron Koch\\Peak Summit Labs\\myronkoch@gmail.com}
\date{April 2026}

\begin{document}
\maketitle

\begin{abstract}
Large language model (LLM) coding agents start each session with an empty context window, discarding accumulated knowledge from prior work. We present PrimeAgentOrchestrator (PAO), a system that spawns new instances of Claude Code -- Anthropic's terminal-based coding agent -- pre-loaded with relevant memories compiled from the user's existing personal databases. At spawn time, PAO queries two independently-operated memory backends in parallel (a PostgreSQL entity-observation database and a Cloudflare Worker semantic search index), fuses results using backend-specific retrieval strategies, and delivers the compiled briefing via filesystem injection that exploits the host agent's configuration auto-read behavior. PAO manages the full agent lifecycle including trust pre-seeding, readiness polling with error detection, and adaptive terminal text injection. We report on four months of regular deployment (December 2025 through March 2026) as an experience report, documenting three generations of context delivery mechanisms, the failure modes that motivated each redesign, and the engineering tradeoffs of bridging heterogeneous memory systems rather than building a unified one.
\end{abstract}

\section{Introduction}

The proliferation of LLM-based coding agents -- Claude Code, GitHub Copilot Workspace, Cursor, Aider -- has shifted the bottleneck in AI-assisted development from model capability to context engineering. A fresh agent session knows nothing about prior work, user preferences, or domain-specific decisions. Users spend the first minutes of each session re-explaining context that existed in a previous session but was discarded when that session ended.

Multi-agent frameworks such as AutoGen~\citep{wu2023autogen}, MetaGPT~\citep{hong2023metagpt}, and CAMEL~\citep{li2023camel} coordinate simultaneously running agents, but each agent starts cold -- initialized only with a task description and system prompt. Memory-augmented systems like MemGPT~\citep{packer2023memgpt} and Mem0~\citep{chhikara2025mem0} provide persistent memory for a single agent across sessions, but do not transfer that memory to newly spawned specialized agents. AgentSpawn~\citep{costa2026agentspawn} transfers memory slices during dynamic spawning, but only within a single task session using intra-process memory, not across sessions using external databases.

This paper presents PrimeAgentOrchestrator (PAO), a system that bridges these concerns. PAO operates as an orchestrator with access to the user's accumulated personal memory, spawning specialized Claude Code agents that inherit relevant slices of that memory at initialization time. The key insight is that personal AI infrastructure accumulates knowledge in multiple independent storage systems -- each with its own query model, schema, and access pattern -- and that this knowledge should be compiled into a coherent briefing and delivered to new agents before they start, not discovered ad-hoc during task execution.

This is an experience report. PAO has been deployed on a single MacBook Pro (Apple M3) for four months (December 2025 through March 2026) as part of a larger personal AI infrastructure. We do not present controlled experiments comparing primed versus cold agents; instead, we document the system's architecture, the failure modes we encountered, and the engineering lessons we learned. The contribution is the architecture itself and the practical knowledge of what works and what breaks when automating terminal-based LLM agent spawning with memory injection.

Our specific contributions are:
\begin{enumerate}
  \item \textbf{Memory-primed agent spawning.} A protocol that queries heterogeneous memory backends in parallel, compiles results into a structured briefing, and delivers it via filesystem injection -- eliminating the timing dependencies of clipboard-based approaches.
  \item \textbf{A ``bridge, don't own'' approach to memory fusion.} Rather than requiring a unified memory architecture, PAO queries pre-existing, independently-operated databases using backend-specific retrieval strategies. We discuss the tradeoffs of this design: lower integration cost and preserved ecosystem independence at the expense of schema coupling and non-uniform retrieval quality.
  \item \textbf{Practical engineering of CLI agent lifecycle management.} Solutions to problems that lack academic treatment but dominate real-world deployment: trust pre-seeding, readiness polling with both positive and negative indicators, adaptive terminal text injection, and concurrent-safe registry management.
\end{enumerate}

\section{Background and Related Work}

\subsection{Multi-Agent LLM Frameworks}

AutoGen~\citep{wu2023autogen} introduced the conversable agent paradigm with structured message-passing. MetaGPT~\citep{hong2023metagpt} imposed software-engineering role hierarchies with a shared message pool. CAMEL~\citep{li2023camel} explored role-playing via inception prompting. AgentVerse~\citep{chen2023agentverse} investigated dynamic team composition and emergent behaviors. All share a fundamental characteristic: agents are initialized with task descriptions and system prompts, not with accumulated personal knowledge from prior sessions. PAO addresses the orthogonal problem of what agents know when they start, not how they coordinate while running.

\subsection{Memory-Augmented Agents}

MemGPT~\citep{packer2023memgpt} introduced virtual context management inspired by OS memory hierarchies, enabling a single agent to maintain persistent memory across sessions via main memory, recall storage, and archival storage. The Generative Agents framework~\citep{park2023generative} demonstrated memory streams with recency-importance-relevance retrieval for simulated social agents. Reflexion~\citep{shinn2023reflexion} showed that verbal self-reflection stored in episodic memory improves agent performance on sequential decision tasks. MAGMA~\citep{jiang2026magma} advanced multi-view memory fusion through four orthogonal relational graphs (semantic, temporal, causal, entity) with intent-aware query routing. Mem0~\citep{chhikara2025mem0} provided a production-scale memory layer with dynamic extraction and graph-based consolidation. \citet{hu2025memory} surveyed the landscape, providing a taxonomy of factual, experiential, and working memory for AI agents.

These systems optimize memory management for a single persistent agent identity. None address transferring accumulated personal memory to newly spawned agents at initialization time. MemGPT's agent manages its own memory during operation; PAO delivers externally-compiled memory once at spawn time. MAGMA fuses multiple memory views within a unified architecture; PAO bridges pre-existing, independently-operated databases without a unifying layer.

\subsection{Agent Spawning with Memory Transfer}

AgentSpawn~\citep{costa2026agentspawn} is the closest prior work. It introduces dynamic agent spawning triggered by runtime complexity metrics during code generation, with a ``SpawnPackage'' that transfers selective memory slices to child agents. AgentSpawn reports improved task completion rates versus static baselines on SWE-bench. The differences from PAO are:

\begin{itemize}
  \item \textbf{Memory scope.} AgentSpawn transfers intra-session parent memory (code files, API documentation, conversation turns). PAO transfers cross-session personal knowledge (semantic facts, episodic memories, entity observations, conversation history accumulated over months).
  \item \textbf{Backend architecture.} AgentSpawn uses a single in-process memory store. PAO queries independently-operated external databases with different query models.
  \item \textbf{Lifecycle management.} AgentSpawn has no persistent agent registry or lifecycle tracking beyond the current task. PAO maintains a concurrent-safe registry, supports pause/resume, checkpointing, and multi-agent coordination commands.
  \item \textbf{Deployment context.} AgentSpawn targets automated code generation benchmarks. PAO targets a human developer's daily workflow across diverse domains.
\end{itemize}

\subsection{Context Engineering for Coding Agents}

\citet{santos2025claude} empirically studied 328 CLAUDE.md configuration files from public Claude Code projects, finding that 72.6\% specify architecture information. This validates CLAUDE.md as the dominant developer-facing context injection mechanism -- the same mechanism PAO exploits for briefing delivery. \citet{bui2026terminal} described the context engineering layer in terminal coding agents, noting that subagents start with fresh contexts -- the opposite of PAO's warm-start philosophy. \citet{zhang2025ace} proposed evolving contexts through generator-reflector-curator roles, optimizing context during task execution rather than at spawn time.

\subsection{Agent Infrastructure}

\citet{chan2025infrastructure} argued for external infrastructure to govern AI agent behavior: identification, interaction shaping, harm detection, and accountability. PAO's agent registry and lifecycle management represent a practical, single-user implementation of the tracking and identification concerns that \citeauthor{chan2025infrastructure} raise at the multi-organization level. The governance implications of PAO's \texttt{--dangerously-skip-permissions} flag -- necessary for automation but granting unrestricted tool access to spawned agents -- are discussed in Section~\ref{sec:security}.

\section{System Architecture}

PAO consists of two TypeScript command-line tools running on the Bun runtime, a shared library of nine modules, and a template file, totaling approximately 1,800 lines at time of initial evaluation (March 2026). The system currently targets Claude Code (Anthropic's terminal-based coding agent) as the host agent; the architecture is specific to Claude Code's configuration system, trust mechanisms, and terminal UI indicators.

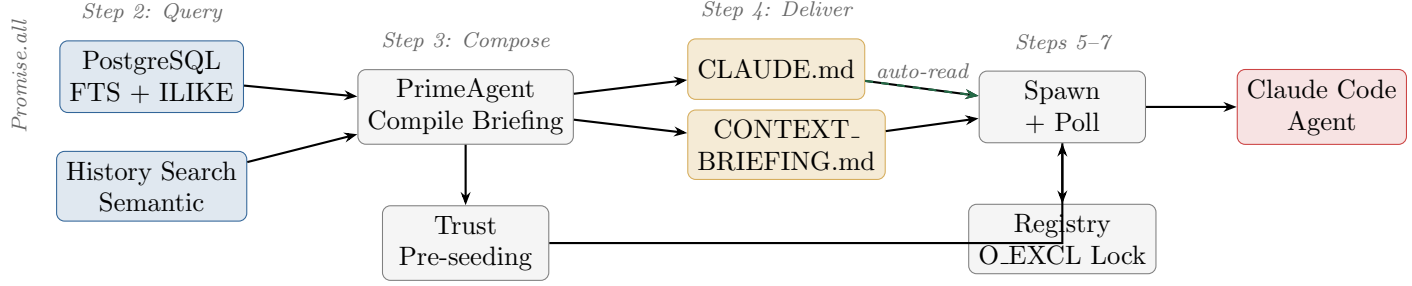
\begin{figure}[t]
\centering
\begin{tikzpicture}[
  node distance=0.8cm and 1.2cm,
  box/.style={rectangle, draw, rounded corners=3pt, minimum width=2.2cm, minimum height=0.8cm, font=\small, align=center},
  backend/.style={box, fill=pgblue!15, draw=pgblue},
  process/.style={box, fill=bglight, draw=black!50},
  file/.style={box, fill=fileyellow!20, draw=fileyellow!80},
  agent/.style={box, fill=agentred!15, draw=agentred},
  arrow/.style={-{Stealth[length=5pt]}, thick},
  label/.style={font=\scriptsize\itshape, text=black!60},
]

\node[backend] (pg) {PostgreSQL\\FTS + ILIKE};
\node[backend, below=0.5cm of pg] (hs) {History Search\\Semantic};

\node[label, left=0.3cm of pg] (par) {\rotatebox{90}{Promise.all}};

\node[process, right=1.5cm of pg, yshift=-0.4cm] (prime) {PrimeAgent\\Compile Briefing};

\node[file, right=1.5cm of prime, yshift=0.5cm] (claude) {CLAUDE.md};
\node[file, right=1.5cm of prime, yshift=-0.5cm] (brief) {CONTEXT\_\\BRIEFING.md};

\node[process, below=0.8cm of prime] (trust) {Trust\\Pre-seeding};

\node[process, right=1.5cm of claude, yshift=-0.5cm] (spawn) {Spawn\\+ Poll};

\node[agent, right=1.2cm of spawn] (agent) {Claude Code\\Agent};

\node[process, below=0.8cm of spawn] (reg) {Registry\\O\_EXCL Lock};

\draw[arrow] (pg) -- (prime);
\draw[arrow] (hs) -- (prime);
\draw[arrow] (prime) -- (claude);
\draw[arrow] (prime) -- (brief);
\draw[arrow] (prime) -- (trust);
\draw[arrow] (trust) -| (spawn);
\draw[arrow] (claude) -- (spawn);
\draw[arrow] (brief) -- (spawn);
\draw[arrow] (spawn) -- (agent);
\draw[arrow] (spawn) -- (reg);

\draw[arrow, dashed, cfgreen!70!black] (claude) -- node[above, label] {auto-read} (spawn);

\node[label, above=0.1cm of pg] {Step 2: Query};
\node[label, above=0.1cm of prime] {Step 3: Compose};
\node[label, above=0.1cm of claude] {Step 4: Deliver};
\node[label, above=0.1cm of spawn] {Steps 5--7};

\end{tikzpicture}
\caption{PAO spawn pipeline. Memory backends are queried in parallel (Step 2), results compiled into a briefing (Step 3), delivered via filesystem injection (Step 4), then the agent is spawned with trust pre-seeding, readiness polling, and registry tracking (Steps 5--7). The dashed arrow indicates Claude Code's automatic reading of CLAUDE.md on startup.}
\label{fig:architecture}
\end{figure}

\subsection{Spawn Pipeline}

When spawning a primed agent, PAO executes a seven-step pipeline (Figure~\ref{fig:architecture}):

\begin{enumerate}
  \item \textbf{Topic extraction and validation.} Natural language topic is parsed, folder paths are auto-detected from the input, agent names are derived from folder basenames, and all inputs are validated against strict patterns (alphanumeric names, no shell metacharacters in paths) before any shell interaction.
  \item \textbf{Parallel memory gathering.} Backends are queried concurrently via \texttt{Promise.all} with per-backend timeouts. Each backend uses its native query model (Section~3.2).
  \item \textbf{Prompt composition.} Retrieved memories are injected into a template with conditional sections. Empty sections are removed. If all backends return empty, the agent receives self-priming instructions telling it to query memory systems in its own session.
  \item \textbf{File-based context delivery.} The briefing is written as \texttt{CONTEXT\_BRIEFING.md} in the agent's working directory. A \texttt{CLAUDE.md} is created (or appended to) with a reference. Claude Code auto-reads \texttt{CLAUDE.md} on startup (Section~3.3).
  \item \textbf{Trust pre-seeding.} Workspace-level and project-level trust artifacts are pre-created (Section~3.4).
  \item \textbf{Spawn.} A new Terminal.app window or tmux pane is opened running \texttt{claude} with the \texttt{-{}-dangerously\hbox{-}skip\hbox{-}permissions} flag.
  \item \textbf{Readiness polling and nudge.} Terminal output is polled for readiness indicators. A lightweight nudge is sent after readiness is confirmed. The nudge is non-critical -- if it fails, the agent still has its context via local files.
\end{enumerate}

\subsection{Heterogeneous Memory Fusion}

PAO queries two backends with fundamentally different query models. This ``bridge, don't own'' approach has a specific motivation: these databases already existed as part of the user's personal AI infrastructure before PAO was built. Rather than migrating all memory into a unified store, PAO queries each backend in its native language. An earlier version of PAO included a third backend (a SQLite-based semantic memory store), which was removed after evaluation revealed that its keyword-based retrieval introduced significant false positives without improving overall precision.

\textbf{Longterm Memory (PostgreSQL).} A normalized entity-observation database (715 observations across 217 entities at time of evaluation). PAO uses PostgreSQL's built-in full-text search (\texttt{to\_tsvector}/\texttt{plainto\_tsquery} with \texttt{ts\_rank}) as the primary retrieval path. When FTS returns no results -- common for proper nouns and domain-specific terms that lack stemming support -- PAO falls back to per-keyword \texttt{ILIKE} substring matching.

\textbf{History Search (Cloudflare Worker).} A semantic search index over past Claude Code conversation transcripts, hosted as a Cloudflare Worker with D1 storage and Vectorize embeddings. PAO communicates via JSON-RPC over HTTP. This backend performs semantic (embedding-based) retrieval rather than lexical matching, providing better recall for conceptually related topics that don't share exact keywords.

\textbf{Tradeoffs of ``bridge, don't own.''} The main advantage is ecosystem independence: each backend evolves independently, serves other consumers, and doesn't require schema migration. The costs are non-uniform retrieval quality, schema coupling, and no cross-backend relevance normalization. We analyze these tradeoffs in detail in Section~\ref{sec:bridging}.

\subsection{File-Based Context Delivery}

The context delivery mechanism evolved through three iterations, each motivated by a specific failure mode:

\begin{table}[h]
\centering
\small
\begin{tabular}{@{}llll@{}}
\toprule
\textbf{Gen.} & \textbf{Mechanism} & \textbf{Failure Mode} & \textbf{Resolution} \\
\midrule
V1 & Write to /tmp, Cmd+V paste & Paste during MCP loading & Added polling \\
V2 & Paste after readiness poll & Early indicators matched welcome screen & Late-stage indicators \\
V3 & Write to working dir, CLAUDE.md ref & None -- timing-independent & Current \\
\bottomrule
\end{tabular}
\caption{Evolution of context delivery mechanisms.}
\label{tab:delivery}
\end{table}

The V3 approach works because Claude Code auto-reads \texttt{CLAUDE.md} from its working directory during startup configuration loading. The briefing files exist before the agent process launches, eliminating race conditions entirely. This auto-read behavior is observed in current versions of Claude Code but is not a documented API guarantee; future versions could change this mechanism without notice. The pattern is validated by \citet{santos2025claude}, who found that 72.6\% of Claude Code projects use CLAUDE.md for architecture and context specification.

We use two files (\texttt{CLAUDE.md} + \texttt{CONTEXT\_BRIEFING.md}) rather than embedding the full briefing in \texttt{CLAUDE.md} because: (a) the target directory may already have a \texttt{CLAUDE.md} with project-specific instructions that should not be overwritten; (b) separating the briefing allows the agent to distinguish its pre-loaded context from project configuration; and (c) the briefing can be regenerated without modifying the project's own \texttt{CLAUDE.md}.

\subsection{Trust Pre-Seeding}

When Claude Code opens a new directory, it may present authorization dialogs that block terminal input. Any text sent to the terminal during these dialogs is silently consumed.

PAO prevents these dialogs by pre-creating trust artifacts at two levels before launching the agent: (1) \texttt{.claude/settings.local.json} in the target directory, containing the user's global MCP server allowlist; and (2) \texttt{\textasciitilde/.claude/projects/\{path-key\}/} where the path-key is derived from the folder path.

\subsection{Readiness Polling}

After spawning the agent process, PAO determines when the agent is ready to accept input by polling terminal output with exponential backoff (1s to 3s intervals, 45s maximum).

\textbf{Positive indicators} (agent is ready): the \texttt{>} prompt character (U+276F), the \texttt{tokens} counter, the \texttt{bypass permissions} mode indicator, and the \texttt{MCPs} server count.

\textbf{Negative indicators} (agent will never become ready): ``No conversation to resume'' (wrong spawn mode), ``Do you trust'' / ``Allow this project'' (trust dialog not suppressed), ``Could not find'' (session error). When a negative indicator is detected, PAO bails immediately with a descriptive error rather than waiting for the 45-second timeout.

\subsection{Terminal Backend Abstraction}

PAO supports two terminal backends through a dispatch layer that routes \texttt{send}, \texttt{read}, \texttt{kill}, and \texttt{isAlive} operations to the appropriate implementation.

\textbf{Terminal.app (macOS only).} Windows are created via AppleScript. Text injection uses clipboard with a separate AppleScript call for the Enter keystroke, with a 500ms delay between paste and submit to prevent the Enter from firing before the input handler has processed the pasted text.

\textbf{tmux (cross-platform).} Panes are created in a dedicated session. Text injection uses \texttt{load-buffer} + \texttt{paste-buffer} (avoiding \texttt{send-keys}, which interprets characters like \texttt{\#} and \texttt{;} as tmux command separators), followed by adaptive polling to confirm the text appeared before sending Enter.

\subsection{Security Considerations}
\label{sec:security}

PAO spawns agents with \texttt{--dangerously-skip-permissions}, granting unrestricted tool access. This is a deliberate tradeoff: the flag is necessary for unattended automation, but it means spawned agents can read, write, and execute anything on the host machine.

This is acceptable in PAO's deployment context -- a single developer on their own machine -- but would be inappropriate for multi-user deployments or shared infrastructure. All user-provided inputs (agent names, folder paths, tmux targets) are validated against strict patterns before reaching any shell command.

\section{Evaluation and Operational Experience}

PAO has been deployed on a MacBook Pro (Apple M3) for four months (December 2025 through March 2026) as part of a personal AI infrastructure with 80+ reusable skills, automated hooks, and cross-machine coordination. We report illustrative case studies comparing primed versus cold agents alongside operational observations from deployment.

\subsection{Illustrative Case Studies: Cold vs. Primed Agents}

To explore whether memory-primed agents produce better initial responses than cold-started agents, we conducted illustrative case studies across five tasks. For each task, we spawned two Claude Code agents simultaneously -- one cold (empty working directory, no briefing) and one primed (with PAO's full pipeline). Both received the same task prompt and were given equal time to respond.

An LLM judge (Claude Haiku 4.5, temperature 0) scored each response on specificity, accuracy, and actionability (1--5 scale each, 15 max). We note that using an Anthropic model to judge Anthropic agent outputs creates a potential same-vendor bias; this limitation is discussed further in Section~5.3.

\begin{table}[h]
\centering
\small
\begin{tabular}{@{}lccl@{}}
\toprule
\textbf{Task} & \textbf{Cold} & \textbf{Primed} & \textbf{Winner} \\
\midrule
Backchannel architecture & 4/15 & 9/15 & Primed \\
MCP server inventory & 11/15 & 8/15 & Cold \\
App store notarization & 3/15 & 10/15 & Primed \\
Cloudflare infrastructure & 12/15 & 10/15 & Cold \\
Agent orchestration & 6/15 & 11/15 & Primed \\
\midrule
\textbf{Average} & \textbf{7.2} & \textbf{9.6} & \textbf{Primed (3/5)} \\
\bottomrule
\end{tabular}
\caption{Cold vs. primed agent comparison (N=5, qualitative case studies).}
\label{tab:eval}
\end{table}

Note: with N=5, these results constitute qualitative case studies, not statistically significant findings. They illustrate directional trends rather than proving a general effect.

Primed agents won 3 of 5 comparisons with an average score of 9.6 versus 7.2 for cold agents. The primed advantage was strongest on tasks requiring domain-specific knowledge. Cold agents won two tasks where aggressive filesystem exploration compensated for missing context.

\subsection{Pipeline Performance}

Both backends are queried in parallel. Average end-to-end pipeline latency across 15 evaluation tasks was 586ms. All 15 evaluation tasks produced briefings successfully (15/15 delivery rate). Average briefing size was 4,816 characters containing 11.5 memory items per task.

\subsection{Failure Modes and Resolutions}

Three categories of spawn failure were identified and resolved during the deployment period:

\textbf{Resume mode on new directories (February 2026).} The spawn command was routed through an interactive mode selector. When called via \texttt{execSync}, stdin was empty, causing non-deterministic mode selection. \textbf{Fix:} Always pass \texttt{--new} explicitly.

\textbf{Trust dialog interception (January 2026).} Authorization dialogs consumed priming prompts before the agent could process them. \textbf{Fix:} Dual-level trust pre-seeding (Section~3.4).

\textbf{Enter key timing on Terminal.app (March 2026).} Large text blocks pasted via clipboard took longer than 200ms to process; the Enter keystroke was swallowed. \textbf{Fix:} Split paste and Enter into separate AppleScript calls with 500ms delay.

\subsection{Backend Evolution}

An earlier version of PAO included a third memory backend (a SQLite-based keyword matching store). During evaluation, an LLM judge (Claude Haiku 4.5, temperature 0) rated each retrieved item as relevant, partially relevant, or irrelevant to the task topic. Precision was computed as (relevant + partially relevant) / total items per task, averaged across 14 in-domain tasks plus 1 out-of-domain control task (15 total). Each backend configuration was evaluated on the same 15 topics with the same judge prompt. Removing the keyword backend reduced false positives on an out-of-domain control task from 12 items to 4, while in-domain precision remained stable (57.4\% with two backends vs.\ 56.9\% with three). The backend was removed, demonstrating a practical benefit of the ``bridge, don't own'' architecture.

\subsection{Test Suite}

PAO includes 124 automated tests across 5 modules covering input validation, output formatting, registry CRUD with O\_EXCL lock serialization, pure logic functions, and monitoring.

\section{Discussion}

\subsection{Bridging vs. Unifying Memory Systems}
\label{sec:bridging}

The ``bridge, don't own'' design was born from pragmatism, not from a principled stance that bridging is always superior to unifying. The memory backends predated PAO and served other consumers. Building a unified memory layer would have required migrating data, maintaining synchronization, and abandoning existing query interfaces. In deployments where no pre-existing memory ecosystem exists, a unified architecture may be preferable.

A related concern is memory contradiction: when backends return conflicting information, PAO has no mechanism for chronological overriding or conflict resolution. Both items appear in the briefing, and resolution is left to the receiving agent's reasoning. This is an inherent limitation of concatenation without joint ranking.

A unified architecture (like MAGMA's multi-graph approach~\citep{jiang2026magma}) would enable cross-backend relevance normalization and joint ranking. The cost would be migration complexity and the loss of ecosystem independence.

\subsection{Generalizability}

PAO is tightly coupled to Claude Code. The file-based delivery exploits CLAUDE.md auto-read; the trust pre-seeding creates Claude Code-specific configuration files; the readiness indicators are Claude Code's specific UI elements. Porting to another coding agent would require reimplementing the delivery mechanism, trust bypass, and readiness detection.

The architectural pattern -- query memory, compile briefing, inject into agent's configuration -- is generalizable. The implementation is not.

\subsection{Limitations of the Evaluation}

Our cold-vs-primed comparison provides initial evidence that priming improves agent responses for knowledge-recall tasks, but the evaluation has significant limitations. The sample size (5 tasks) is too small for statistical significance. The LLM judge may have biases. The cold agents had access to the same MCP servers and could discover the same information through tool use -- the priming advantage may diminish over multi-turn interactions.

\section{Limitations}

\textbf{Single-user, single-agent-platform design.} PAO is personal infrastructure for one developer targeting one coding agent.

\textbf{Small-scale evaluation.} Our cold-vs-primed comparison covers 5 tasks with an LLM judge. A larger study with human judges would strengthen the findings.

\textbf{Schema coupling.} The query strategies are specific to each backend's current schema.

\textbf{Limited retrieval quality assessment.} Section 4.4 reports a comparative precision evaluation that informed the removal of one backend, but the remaining backends lack per-item confidence scoring or relevance filtering beyond their native ranking.

\textbf{macOS dependency.} The Terminal.app backend requires AppleScript and Accessibility permissions.

\section{Conclusion}

PrimeAgentOrchestrator addresses the cold-start problem for LLM coding agents at the infrastructure level, without modifying the agents themselves. By querying heterogeneous personal memory backends at spawn time and delivering compiled briefings via the host agent's configuration auto-read mechanism, PAO enables new agents to start with relevant cross-session knowledge.

The practical engineering contributions -- trust pre-seeding, readiness polling with error detection, adaptive terminal injection -- address failure modes that receive little attention in the academic literature but arise repeatedly in practice when automating terminal-based LLM agents. The ``bridge, don't own'' approach to memory fusion offers a pragmatic alternative to unified memory architectures for users whose knowledge already spans multiple independent systems.

The primary limitation is the small scale of the evaluation (N=5 case studies with a same-vendor LLM judge). Future work includes larger controlled experiments with cross-vendor judges, memory quality feedback loops, and generalization beyond Claude Code to other coding agent platforms.

\bibliographystyle{plainnat}
\bibliography{references}

@article{wu2023autogen,
  title={AutoGen: Enabling Next-Gen LLM Applications via Multi-Agent Conversation},
  author={Wu, Qingyun and Bansal, Gagan and Zhang, Jieyu and Wu, Yiran and Li, Beibin and Zhu, Erkang and Jiang, Li and Zhang, Xiaoyun and Zhang, Shaokun and Liu, Jiale and Awadallah, Ahmed Hassan and White, Ryen W. and Burger, Doug and Wang, Chi},
  journal={arXiv preprint arXiv:2308.08155},
  year={2023}
}

@article{packer2023memgpt,
  title={MemGPT: Towards LLMs as Operating Systems},
  author={Packer, Charles and Wooders, Sarah and Lin, Kevin and Fang, Vivian and Patil, Shishir G. and Stoica, Ion and Gonzalez, Joseph E.},
  journal={arXiv preprint arXiv:2310.08560},
  year={2023}
}

@article{park2023generative,
  title={Generative Agents: Interactive Simulacra of Human Behavior},
  author={Park, Joon Sung and O'Brien, Joseph C. and Cai, Carrie J. and Morris, Meredith Ringel and Liang, Percy and Bernstein, Michael S.},
  journal={arXiv preprint arXiv:2304.03442},
  year={2023}
}

@article{shinn2023reflexion,
  title={Reflexion: Language Agents with Verbal Reinforcement Learning},
  author={Shinn, Noah and Cassano, Federico and Berman, Edward and Gopinath, Ashwin and Narasimhan, Karthik and Yao, Shunyu},
  journal={arXiv preprint arXiv:2303.11366},
  year={2023}
}

@article{costa2026agentspawn,
  title={AgentSpawn: Adaptive Multi-Agent Collaboration Through Dynamic Spawning for Long-Horizon Code Generation},
  author={Costa, Igor},
  journal={arXiv preprint arXiv:2602.07072},
  year={2026}
}

@article{jiang2026magma,
  title={MAGMA: A Multi-Graph based Agentic Memory Architecture for AI Agents},
  author={Jiang, Dongming and Li, Yi and Li, Guanpeng and Li, Bingzhe},
  journal={arXiv preprint arXiv:2601.03236},
  year={2026}
}

@article{chhikara2025mem0,
  title={Mem0: Building Production-Ready AI Agents with Scalable Long-Term Memory},
  author={Chhikara, Prateek and Khant, Dev and Aryan, Saket and Singh, Taranjeet and Yadav, Deshraj},
  journal={arXiv preprint arXiv:2504.19413},
  year={2025}
}

@article{zhang2025ace,
  title={Agentic Context Engineering: Evolving Contexts for Self-Improving Language Models},
  author={Zhang, Qizheng and Hu, Changran and Upasani, Shubhangi and Ma, Boyuan and Hong, Fenglu and Kamanuru, Vamsidhar and Rainton, Jay and Wu, Chen and Ji, Mengmeng and Li, Hanchen and Thakker, Urmish and Zou, James and Olukotun, Kunle},
  journal={arXiv preprint arXiv:2510.04618},
  year={2025}
}

@article{santos2025claude,
  title={Decoding the Configuration of AI Coding Agents: Insights from Claude Code Projects},
  author={Santos, Helio Victor F. and Costa, Vitor and Montandon, Joao Eduardo and Valente, Marco Tulio},
  journal={arXiv preprint arXiv:2511.09268},
  year={2025}
}

@article{bui2026terminal,
  title={Building Effective AI Coding Agents for the Terminal: Scaffolding, Harness, Context Engineering, and Lessons Learned},
  author={Bui, Nghi D. Q.},
  journal={arXiv preprint arXiv:2603.05344},
  year={2026}
}

@article{chan2025infrastructure,
  title={Infrastructure for AI Agents},
  author={Chan, Alan and Wei, Kevin and Huang, Sihao and Rajkumar, Nitarshan and Perrier, Elija and Lazar, Seth and Hadfield, Gillian K. and Anderljung, Markus},
  journal={arXiv preprint arXiv:2501.10114},
  year={2025}
}

@article{hu2025memory,
  title={Memory in the Age of AI Agents},
  author={Hu, Yuyang and Liu, Shichun and Yue, Yanwei and Zhang, Guibin and Liu, Boyang and Zhu, Fangyi and Lin, Jiahang and Guo, Honglin and Dou, Shihan and Xi, Zhiheng and others},
  journal={arXiv preprint arXiv:2512.13564},
  year={2025}
}

@article{hong2023metagpt,
  title={MetaGPT: Meta Programming for A Multi-Agent Collaborative Framework},
  author={Hong, Sirui and Zhuge, Mingchen and Chen, Jiaqi and Zheng, Xiawu and Cheng, Yuheng and Zhang, Ceyao and Wang, Jinlin and Wang, Zili and Yau, Steven Ka Shing and Lin, Zijuan and Zhou, Liyang and Ran, Chenyu and Xiao, Lingfeng and Wu, Chenglin and Schmidhuber, J{\"u}rgen},
  journal={arXiv preprint arXiv:2308.00352},
  year={2023}
}

@article{li2023camel,
  title={CAMEL: Communicative Agents for ``Mind'' Exploration of Large Language Model Society},
  author={Li, Guohao and Hammoud, Hasan Abed Al Kader and Itani, Hani and Khizbullin, Dmitrii and Ghanem, Bernard},
  journal={arXiv preprint arXiv:2303.17760},
  year={2023}
}

@article{chen2023agentverse,
  title={AgentVerse: Facilitating Multi-Agent Collaboration and Exploring Emergent Behaviors},
  author={Chen, Weize and Su, Yusheng and Zuo, Jingwei and Yang, Cheng and Yuan, Chenfei and Chan, Chi-Min and Yu, Heyang and Lu, Yaxi and Hung, Yi-Hsin and Qian, Chen and Qin, Yujia and Cong, Xin and Xie, Ruobing and Liu, Zhiyuan and Sun, Maosong and Zhou, Jie},
  journal={arXiv preprint arXiv:2308.10848},
  year={2023}
}

\end{document}